\documentclass[letterpaper, 10 pt, conference]{ieeeconf}  
 \IEEEoverridecommandlockouts      

\usepackage{multirow}
\usepackage{cite}
\usepackage{amsmath,amssymb,amsfonts}
\usepackage{algorithmic}
\usepackage{graphicx}
\usepackage{textcomp}
\usepackage{xcolor}
\usepackage{comment} 
\usepackage{colortbl}  
\usepackage{booktabs}
\usepackage{tikz} 
\usepackage[margin=0.8in]{geometry}
\def\BibTeX{{\rm B\kern-.05em{\sc i\kern-.025em b}\kern-.08em
    T\kern-.1667em\lower.7ex\hbox{E}\kern-.125emX}}
\begin{document}

\title{One Step Closer: Creating the Future to Boost Monocular Semantic Scene Completion}
\author{Haoang Lu$^{1}$, Yuanqi Su$^{1 \dag}$,  Xiaoning Zhang$^{1}$ and Hao Hu$^{2}$
\thanks{ \dag Corresponding Author. Email: yuanqisu@mail.xjtu.edu.cn}
\thanks{The research was supported by the Science and Technology Research and Development Plan of China State Railway Group Co., Ltd. (No. RITS2023KF03) and the National Natural Science Foundation of China (No. 62473306, No. U24B20181).}
\thanks{$^{1}$Haoang Lu, Yuanqi Su and Xiaoning Zhang are with School of Computer Science and Technology, Xi'an Jiaotong University, Xi'an, Shaanxi, China
        {\tt\small Orcid: 0009-0008-4514-1110, 0000-0001-7520-7020, 0000-0002-0583-4616}}%
\thanks{$^{2}$Hao Hu is with The Institute of Computing, China Academy of Railway Sciences Corporation and The Center of National Railway Intelligent Transportation System Engineering and Technology
        {\tt\small Orcid: 0000-0002-9422-750X}}%
} 

\newcommand{\colorboxsquare}[2]{%
\ensuremath{%
    \stackrel{\rotatebox{90}{\footnotesize #2}}{%
        \tikz[baseline=(char.base), xshift=1mm]\node[fill=#1, minimum width=2.5mm, minimum height=2.5mm, inner sep=0pt] (char) {};%
    }
    }
}

\maketitle

\begin{abstract}
In recent years, visual 3D Semantic Scene Completion (SSC) has emerged as a critical perception task for autonomous driving due to its ability to infer complete 3D scene layouts and semantics from single 2D images. However, in real-world traffic scenarios, a significant portion of the scene remains occluded or outside the camera's field of view - a fundamental challenge that existing monocular SSC methods fail to address adequately.

To overcome these limitations, we propose Creating the Future SSC (CF-SSC), a novel temporal SSC framework that leverages pseudo-future frame prediction to expand the model‘s effective perceptual range. Our approach combines poses and depths to establish accurate 3D correspondences, enabling geometrically-consistent fusion of past, present, and predicted future frames in 3D space. Unlike conventional methods that rely on simple feature stacking, our 3D-aware architecture achieves more robust scene completion by explicitly modeling spatial-temporal relationships. 

 Comprehensive experiments on SemanticKITTI\cite{behley2019semantickitti} and SSCBench-KITTI-360\cite{li2024sscbench,liao2022kitti} benchmarks demonstrate state-of-the-art performance, validating the effectiveness of our approach, highlighting our method's ability to improve occlusion reasoning and 3D scene completion accuracy.
\end{abstract}

\section{Introduction}
The rapid development of intelligent transportation systems demands robust environmental perception capabilities, particularly in complex urban scenarios where occlusions and limited sensor coverage pose significant challenges. As a fundamental task for autonomous driving and smart city infrastructure, 3D Semantic Scene Completion (SSC) bridges the critical gap between raw sensor inputs and comprehensive 3D environmental understanding. Traditional approaches relying on LiDAR~\cite{song2017semantic,roldao2020lmscnet,yan2021sparse} or multi-view systems~\cite{li2023bridging,li2024hierarchical,huang2023tri} face inherent limitations in cost-effectiveness and deployment scalability, making monocular SSC an increasingly vital solution for next-generation transportation intelligence.
\begin{figure}[htbp]
  \centering
  \includegraphics[width=1\linewidth]{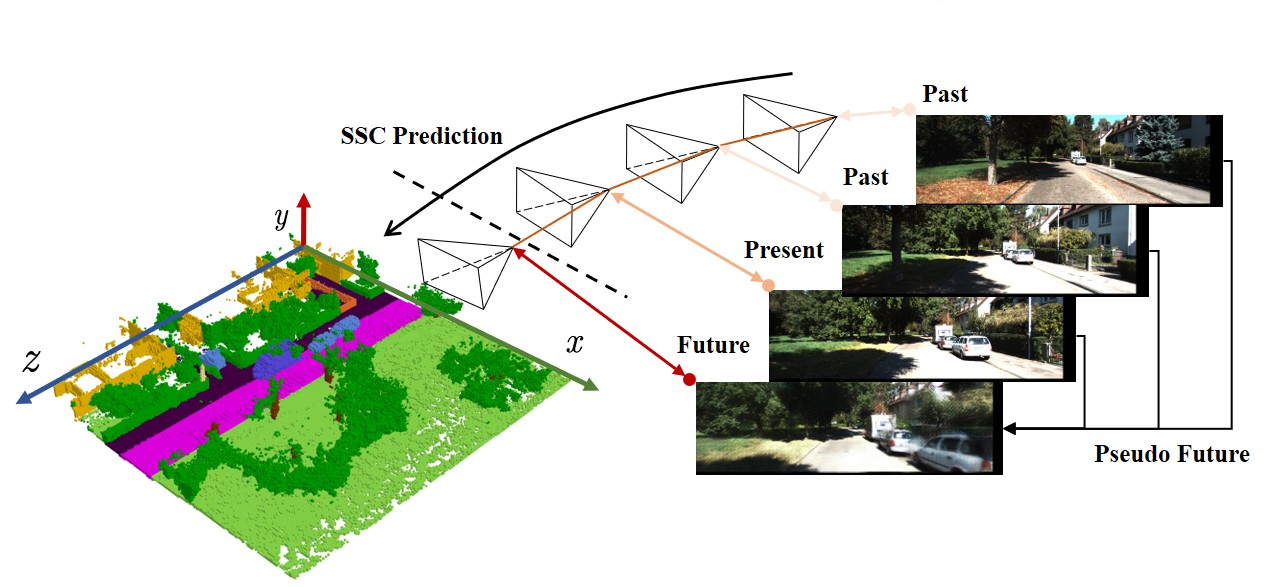}
   \caption{The motivation of our approach: Predicting pseudo future frames from past frames to achieve "seeing ahead" and monocular SSC.}
   \label{fig:1}
\end{figure}

However, beyond its inherently ill-posed nature, monocular SSC faces additional practical constraints from the limited field-of-view (FoV) of single cameras. This severe visibility constraint fundamentally limits the reliability of conventional monocular SSC systems~\cite{cao2022monoscene,yao2023ndc,jiang2024symphonize}, which process each frame in isolation. While introducing temporal information from past frames could theoretically expand the observable range, in practice these frames (captured from positions behind the current viewpoint) provides limited benefits for forward occlusions in the driving direction. To truly extend the perceptual range of monocular SSC we need to "get one step closer" by leveraging future frames.
\begin{figure*}[htbp]
  \centering
  \includegraphics[width=0.9\linewidth]{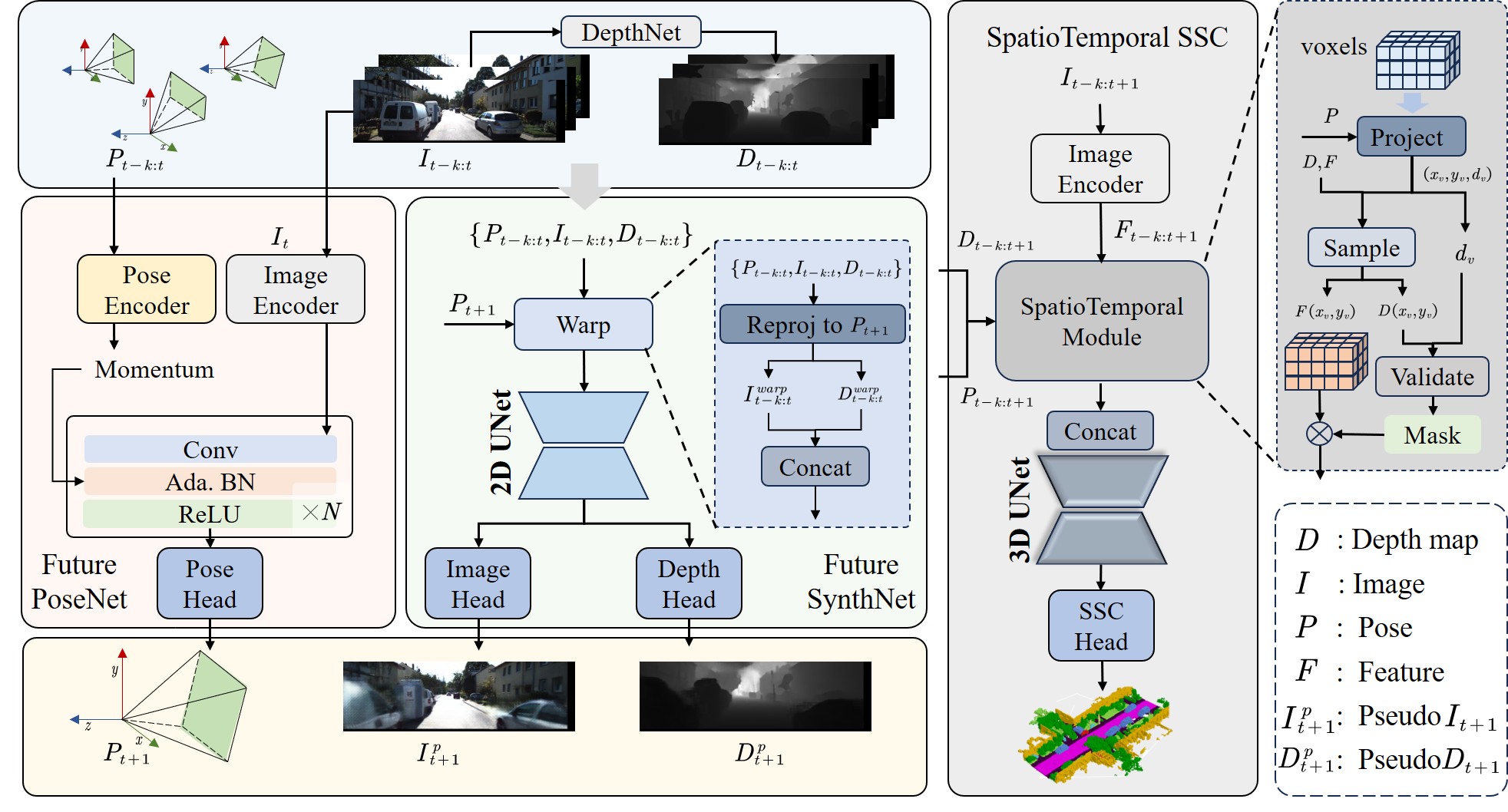}
   \caption{Overall framework of our proposed CF-SSC, which takes images and poses of past frames and the current frame as input, predicts images, depth, and relative poses of future frames, and integrates all temporal information for SSC.}
   \label{fig:method}
\end{figure*}

Therefore, in this paper,we propose a novel framework that utilizes pseudo-future frame prediction to effectively “see ahead” in the scene. Following previous monocular SSC methods\cite{li2023voxformer,jiang2024symphonize,wang2024h2gformer}, we utilize a pre-trained depth prediction model to estimate depth maps for past and current frames. Based on the predicted depth maps and relative poses, we derive optical flow and use it to generate preliminary future frame images and depth maps, which are further refined using a FutureSynthNet to obtain high-quality pseudo-images and pseudo-depth maps of future frames. Finally, by integrating the predicted depth maps and poses, we consolidate all frame information into a unified 3D space for Semantic Scene Completion (SSC).

In contrast to previous temporal monocular SSC methods\cite{li2023voxformer,wang2024h2gformer,zheng2024monoocc} that naively concatenate multi-frame inputs and consequently suffer from geometric ambiguity, our approach achieves temporally coherent 3D fusion through integration of sequential context. Furthermore, by incorporating future frame prediction, we significantly extend the visible scope of SSC, enabling the system to truly "see ahead" and anticipate occluded or emerging structures, shown in Fig.\ref{fig:1}.

Based on the aforementioned modules and methodologies, we present Creating the Future SSC(CF-SSC), an innovative monocular SSC framework that effectively combines 3D geometric reconstruction with future frame prediction. We have conducted extensive experiments on two established benchmark datasets, SemanticKITTI\cite{behley2019semantickitti} and SSCBench-KITTI-360\cite{li2024sscbench,liao2022kitti}, to thoroughly validate our approach, which achieves state-of-the-art performance. The key contributions of our work are summarized as follows:
\begin{itemize}
\item We propose a novel pseudo-future based monocular SSC framework that effectively extends the visible range of monocular SSC through future frame image and depth map prediction. This innovative approach achieves the "see ahead" capability, successfully addressing the limited visible range problem inherent in conventional monocular SSC methods.
\item By combining relative poses with depth prediction, we develop a geometrically consistent temporal monocular SSC method. This 3D-aware fusion effectively eliminates the geometric ambiguity present in previous temporal approaches\cite{li2023voxformer,wang2024h2gformer,zheng2024monoocc}.
\item We conducted extensive experiments on both SemanticKITTI\cite{song2017semantic} and SSCBench-KITTI-360\cite{li2024sscbench,liao2022kitti} datasets to validate our TempSSC framework. The results demonstrate its effectiveness and show state-of-the-art performance.
\end{itemize}
\section{Related Work}
\subsection{Semantic Scene Completion}
3D Semantic Scene Completion (SSC) aims to reconstruct a voxelized 3D scene by predicting both occupancy (geometric completeness) and semantic labels for all voxels within a predefined volume. This task was first formally introduced by\cite{song2017semantic}, establishing a unified framework for joint geometric and semantic inference. Due to its ability to provide dense 3D scene layouts—critical for path planning and obstacle avoidance—SSC has emerged as a pivotal perception task in autonomous driving and robotics.

Early SSC methods\cite{roldao2020lmscnet,li2020anisotropic,yan2021sparse,cheng2021s3cnet} predominantly relied on dense depth input (e.g., LiDAR point clouds or depth maps) to achieve high-fidelity 3D reconstruction.While demonstrating impressive results, their dependence on expensive sensors significantly limited practical deployment scalability.
\begin{figure*}[htbp]
  \centering
  \includegraphics[width=0.9\linewidth]{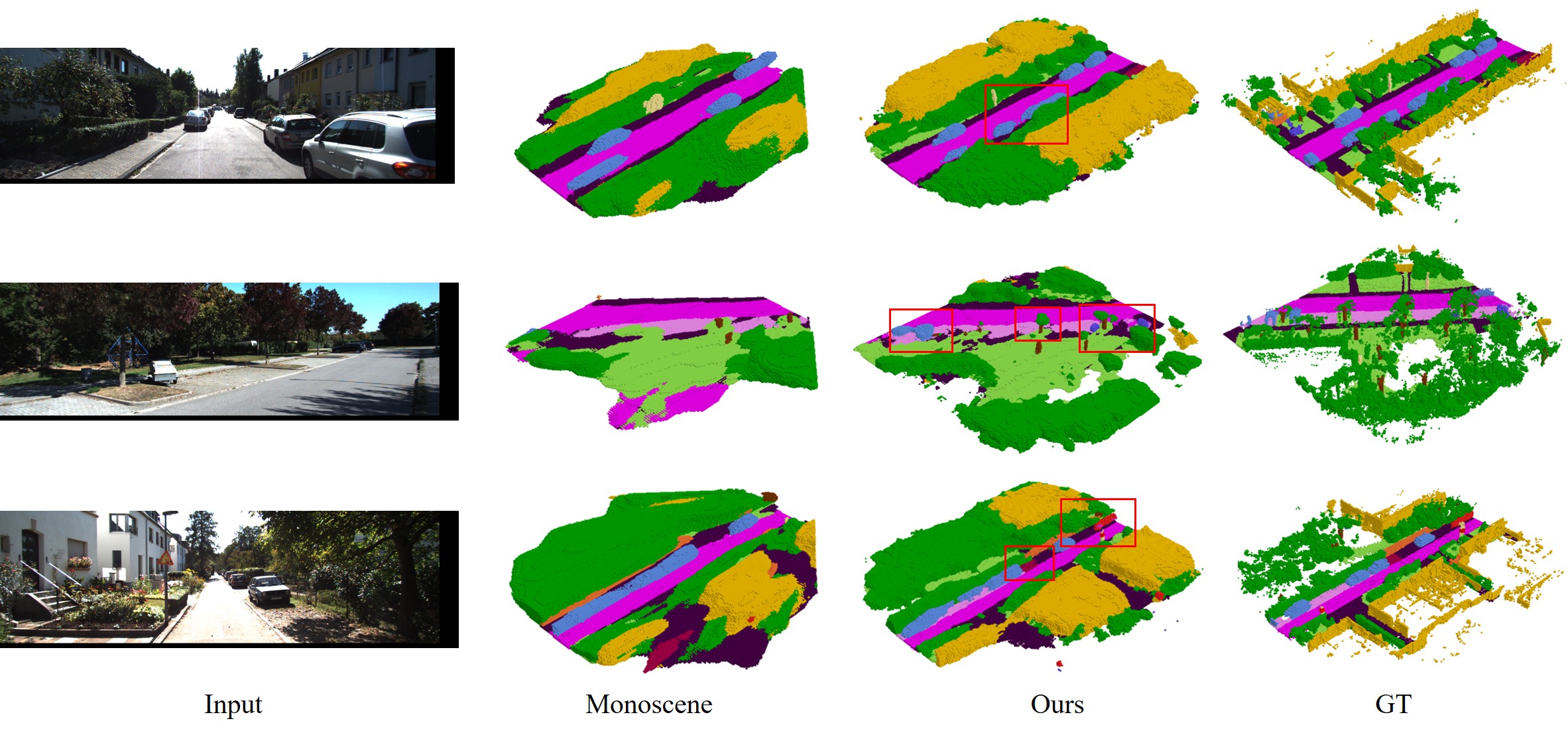}
   \caption{Visualization results of our method on the SemanticKITTI\cite{behley2019semantickitti} dataset along with comparisons against MonoScene\cite{cao2022monoscene}.}
   \label{fig:vis}
\end{figure*}

Recent years have witnessed the rise of vision-based SSC approaches, attracting growing research interest. Notably, \cite{li2024bevformer} proposed a versatile framework applicable to various 3D vision tasks, while \cite{huang2023tri} introduced an efficient tri-plane 3D feature representation that reduces memory requirements. For temporal information fusion, \cite{ye2024cvt} developed a novel cost volume approach, and \cite{li2024hierarchical} proposed Hierarchical Temporal Context Learning. Although these stereo/multi-camera based methods offer lower deployment costs than LiDAR-dependent solutions, they still face inherent limitations including multi-camera system calibration challenges and incompatibility with monocular image inputs.

\subsection{Monocular Semantic Scene Completion}
The field witnessed a significant advancement when \cite{cao2022monoscene} first proposed a purely vision-based monocular SSC solution, which sampled image features by projecting voxel positions onto 2D planes. This work was subsequently improved by \cite{yao2023ndc} through the introduction of Normalized Device Coordinates (NDC) to optimize the 2D-to-3D lifting scheme.

A crucial development came with \cite{li2023voxformer}, who pioneered the integration of depth prediction with monocular SSC. Their approach utilized a pre-trained depth predictor to generate depth maps, enabling active projection of image features into 3D space. This design paradigm has been consistently maintained in subsequent works \cite{jiang2024symphonize,wang2024h2gformer,wang2024not,zheng2024monoocc} (including our depth predictor implementation), with some studies introducing incremental improvements to this framework.

Despite progress in monocular SSC, the limited field of view remains a core challenge. We address this by leveraging future frame prediction to extend perception beyond the current observation.

\subsection{Future Frame Prediction}

Future frame prediction aims to forecast upcoming frames based on past frame information. Depending on the prediction objectives, it can generally be categorized into optical flow-based methods\cite{wu2022optimizing,hu2023dynamic} and direct image generation methods\cite{choi2020channel,lu2022video}. These approaches can be further divided into deterministic synthesis\cite{villar2022mspred,park2023biformer} and stochastic synthesis\cite{franceschi2020stochastic,wu2021greedy} based on their generation mechanisms.

As our goal is accurate future frame prediction rather than realism, we adopt deterministic synthesis. Focusing on occluded regions, which challenge optical flow assumptions, we use direct image generation. To meet real-time demands in traffic scenarios, we also avoid diffusion-based models\cite{ho2020denoising,rombach2022high,blattmann2023align}.


\section{Method}
\subsection{Overview}
As shown in Fig.\ref{fig:method}, our method first feeds the relative pose sequence of past frames $P_{t-k:t}$ and the current frame $I_{t}$ into the FuturePoseNet to predict future frame poses $P_{t+1}^{p}$. Next, we compute the optical flow between future frames and both current and past frames using the predicted depth maps and relative poses, generating initial pseudo-future frames and their corresponding depth. These are then refined by the FutureSynthNet to produce the final predicted future frames and depth maps.

Finally, all frames, along with their depth maps and poses, are fed into the SpatioTemporal SSC, where image features are projected into a unified 3D space for geometrically consistent temporal integration.

\begin{figure}[htbp]
\centerline{\includegraphics[width=1\columnwidth]{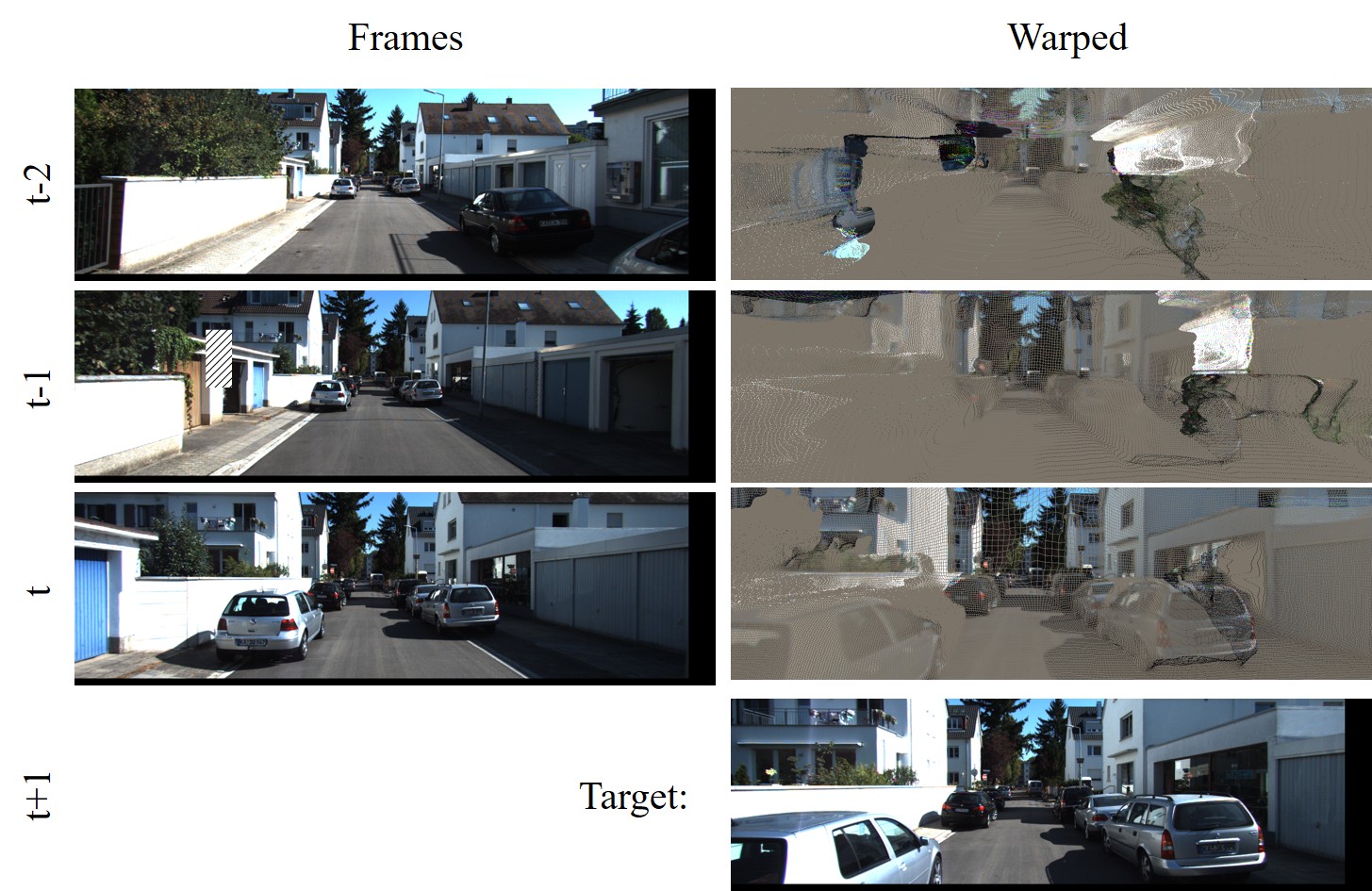}}
\caption{Visualization of warping past and current frames to future frame.}
\label{fig:warp}
\end{figure}
\definecolor{road}{RGB}{255, 0, 255}
\definecolor{sidewalk}{RGB}{75, 0, 75}
\definecolor{parking}{RGB}{255, 150, 255}
\definecolor{other-grnd.}{RGB}{175, 0, 75}
\definecolor{building}{RGB}{255, 200, 0}
\definecolor{car}{RGB}{100, 150, 245}
\definecolor{truck}{RGB}{80, 30, 180}
\definecolor{bicycle}{RGB}{100, 230, 245}
\definecolor{motorcycle}{RGB}{30, 60, 150}
\definecolor{other-veh.}{RGB}{100, 80, 250}
\definecolor{vegetation}{RGB}{0, 175, 0}
\definecolor{trunk}{RGB}{135, 60, 0}
\definecolor{terrain}{RGB}{150, 240, 80}
\definecolor{person}{RGB}{255, 30, 30}
\definecolor{bicyclist}{RGB}{255, 40, 200}
\definecolor{motorcyclist}{RGB}{150, 30, 90}
\definecolor{fence}{RGB}{255, 120, 50}
\definecolor{pole}{RGB}{255, 240, 150}
\definecolor{traf.-sign}{RGB}{255, 0, 0}

\begin{table*}[htbp]
\caption{ Quantitative results on the hidden test set of SemanticKITTI~\cite{behley2019semantickitti}, where the highest and second-highest scores for each metric are highlighted in \textbf{bold} and \underline{underline}, respectively.} 
    \footnotesize
    \setlength{\tabcolsep}{1pt}
    \renewcommand{\arraystretch}{1.2}
    \centering 
    \begin{tabular}{c|c|cc|ccccccccccccccccccc}
        \toprule
        \textbf{Method} & \textbf{Date} & \textbf{IoU$\uparrow$} & \textbf{mIoU$\uparrow$} & \colorboxsquare{road}{road} & \colorboxsquare{sidewalk}{sidewalk} & \colorboxsquare{parking}{parking} & \colorboxsquare{other-grnd.}{other-grnd.} & \colorboxsquare{building}{building} & \colorboxsquare{car}{car} & \colorboxsquare{truck}{truck} & \colorboxsquare{bicycle}{bicycle} & \colorboxsquare{motorcycle}{motorcycle} & \colorboxsquare{other-veh.}{other-veh.} & \colorboxsquare{vegetation}{vegetation} & \colorboxsquare{trunk}{trunk} & \colorboxsquare{terrain}{terrain} & \colorboxsquare{person}{person} & \colorboxsquare{bicyclist}{bicyclist} & \colorboxsquare{motorcyclist}{motorcyclist} & \colorboxsquare{fence}{fence} & \colorboxsquare{pole}{pole} & \colorboxsquare{traf.-sign}{traf.-sign} \\ 
        \midrule
        \multicolumn{22}{@{}l}{\quad\textbf{Stereo camera-based methods}} \\ \hline
        StereoScene\cite{li2023bridging} & IJCAI2024 & 43.34 & 15.36 & 61.90 & 31.20 & 30.70 & 10.70 & 24.20 & 22.80 & 2.80 & 3.40 & 2.40 & 6.10 & 23.80 & 8.40 & 27.00 & 2.90 & 2.20 & 0.50 & 16.50 & 7.00 & 7.20 \\
        HTCL-S\cite{li2024hierarchical} & ECCV2024 & 44.23 & 17.09 & 64.40 & 34.80 & 33.80 & 12.40 & 25.90 & 27.30 & 5.70 & 1.80 & 2.20 & 5.40 & 25.30 & 10.80 & 31.20 & 1.10 & 3.10 & 0.90 & 21.10 & 9.00 & 8.30 \\
        \hline
        \multicolumn{22}{@{}l}{\quad\textbf{Monocular camera-based methods}} \\ \hline
        MonoScene\cite{cao2022monoscene}& CVPR2023 & 34.16 & 11.08 & 54.70 & 27.10 & 24.80 & 5.70 & 14.40 & 18.80 & 3.30 & 0.50 & 0.70 & 4.40 & 14.90 & 2.40 & 19.50 & 1.00 & 1.40 & 0.40 & 11.10 & 3.30 & 2.10 \\
        TPVFormer\cite{huang2023tri} & CVPR2023 & 34.25 & 11.26 & 55.10 & 27.20 & 27.40 & 6.50 & 14.80 & 19.20 & 3.70 & 1.00 & 0.50 & 2.30 & 13.90 & 2.60 & 20.40 & 1.10 & 2.40 & 0.30 & 11.00 & 2.90 & 1.50 \\
        SurroundOcc\cite{wei2023surroundocc} & ICCV2023 & 34.72 & 11.86 & 56.90 & 28.30 & \textbf{30.20} & 6.80 & 15.20 & 20.60 & 1.40 & 1.60 & 1.20 & 4.40 & 14.90 & 3.40 & 19.30 & 1.40 & 2.00 & 0.10 & 11.30 & 3.90 & 2.40 \\
        OccFormer\cite{zhang2023occformer} & ICCV2023 & 34.53 & 12.32 & 55.90 & 30.30 & 31.50 & 6.50 & 15.70 & 21.60 & 1.20 & 1.50 & 1.70 & 3.20 & 16.80 & 3.90 & 21.30 & 2.20 & 1.10 & 0.20 & 11.90 & 3.80 & 3.70 \\
        IAMSSC\cite{xiao2024instance} & T-ITS2024 & 43.74 & 12.37 & 54.00 & 25.50 & 24.70 & 6.90 & 19.20 & 21.30 & 3.80 & 1.10 & 0.60 & 3.90 & 22.70 & 5.80 & 19.40 & 1.50 & 2.90 & 0.50 & 11.90 & 5.30 & 4.10 \\
        VoxFormer-T\cite{li2023voxformer} & CVPR2023 & 43.21 & 13.41 & 54.10 & 26.90 & 25.10 & 7.30 & 23.50 & 21.70 & 3.60 & 1.90 & 1.60 & 4.10 & 24.40 & 8.10 & 24.20 & 1.60 & 1.10 & 0.00 & 13.10 & 6.60 & 5.70 \\
        DepthSSC\cite{yao2023depthssc} & arXiV2024 & \underline{44.58} & 13.11 & 55.64 & 27.25 & 25.72 & 5.78 & 20.46 & 21.94 & 3.74 & 1.35 & 0.98 & 4.17 & 23.37 & 7.64 & 21.56 & 1.34 & 2.79 & 0.28 & 12.94 & 5.87 & 6.23 \\
        Symphonize\cite{jiang2024symphonize} & CVPR2024 & 42.19 & 15.04 & 58.40 & 29.30 & 26.90 & \underline{11.70} & \underline{24.70} & 23.60 & 3.20 & \underline{3.60} & \underline{2.60} & 5.60 & 24.20 & 10.00 & 23.10 & \textbf{3.20} & 1.90 & \textbf{2.00} & 16.10 & \textbf{7.70} & 8.00 \\
        HASSC-T\cite{wang2024not} & CVPR2024 & 42.87 & 14.38 & 55.30 & 29.60 & 25.90 & 11.30 & 23.10 & 23.00 & 2.90 & 1.90 & 1.50 & 4.90 & 24.80 & 9.80 & \underline{26.50} & 1.40 & \underline{3.00} & 0.00 & 14.30 & 7.00 & 7.10 \\
        H2GFormer-T\cite{wang2024h2gformer} & AAAI2024 & 43.52 & 14.60 & 57.90 & 30.40 & 30.00 & 6.90 & 24.00 & 23.70 & \underline{5.20} & 0.60 & 1.20 & 5.00 & \underline{25.20} & \underline{10.70} & 25.80 & 1.10 & 0.10 & 0.00 & 14.60 & \underline{7.50} & \textbf{9.30} \\
        MonoOcc-L\cite{zheng2024monoocc}& ICRA2024 & - & \underline{15.63} & \underline{59.10} & \underline{30.90} & 27.10 & 9.80 & 22.90 & \underline{23.90} & \textbf{7.20} & \textbf{4.50} & 2.40 & \textbf{7.70} & 25.00 & 9.80 & 26.10 & \underline{2.80} & \textbf{4.70} & \underline{0.60} & \underline{16.90} & 7.30 & 8.40 \\
        \hline
        CF-SSC-Online & - & \textbf{46.21} & \textbf{16.40} & \textbf{61.30} & \textbf{33.30} & \underline{29.20} & \textbf{11.90} & \textbf{30.40} & \textbf{26.30} & 4.80 & 2.60 & \textbf{2.70} & \underline{6.30} & \textbf{28.50} & \textbf{11.40} & \textbf{28.30} & 1.50 & 1.40 & 0.40 & \textbf{17.70} & 7.20 & 6.30\\
        \hline
        CF-SSC-Offline & - & 48.25 & 17.70 & 62.20 & 34.80 & 30.70 & 12.90 & 31.70 & 27.70 & 6.00 & 3.50 & 3.50 & 6.20 & 31.60 & 12.90 & 29.90 & 2.20 & 2.80 & 0.80 & 19.10 & 9.10 & 8.70\\
        \bottomrule
    \end{tabular}
    \label{tab:smkitti}
\end{table*}
\subsection{FuturePoseNet}
To predict the direction of the next frame's pose, we need two pieces of information: the current vehicle's momentum and the current road information, which we obtain from the pose sequence of past frames $P_{t-k:t}$ and the current frame's image $I_{t}$, respectively. As shown in Fig.\ref{fig:method}, we first use the current frame's image feature into 3D, using the  same method as in SpatioTemporal SSC (but using only a single frame) and obtain the 3D feature. Next, we encode the pose sequence of past frames $P_{t-k:t}$ to derive the vehicle's momentum feature $M_{t}$. Finally, we employ a 3D U-Net network, replacing all batch normalization \cite{ioffe2015batch} layers with adaptive batch normalization\cite{huang2017arbitrary}, and use the vehicle's momentum feature $M_{t}$ to modulate the 3D feature, ultimately predicting the next frame's pose $P_{t+1}^p$.

This approach integrates vehicle dynamics with 3D scene understanding for robust future pose prediction, using adaptive 3D feature modulation to ensure accurate motion forecasting and geometric consistency.

\subsection{FutureSynthNet}
By leveraging predicted depth maps of current and past frames $D_{t-k:t}$, along with relative poses between past-current frames and future-current frames (predicted by FuturePoseNet) $P_{t-k:t+1}$, we reproject past and current images onto the future frame. This allows us to generate pseudo-future frames and corresponding depth maps through warping. As shown in the Fig.\ref{fig:warp}, the generation quality degrades for frames temporally distant from the target future frame due to object motion. 

These pseudo-future results are then refined by FutureSynthNet (a U-Net based network) to produce optimized predictions. Notably, during FutureSynthNet training, we preserve 3D consistency by employing both L1 loss against ground truth and a feature-level loss computed using SpatioTemporal SSC's backbone network.

\subsection{SpatioTemporal SSC}
As shown in Fig.\ref{fig:method}, SpatioTemporal SSC consists of three components: a 2D backbone, a spatiotemporal module, and a 3D U-Net. The 2D backbone extracts 2D features from each frame's image, while the spatiotemporal module spatially aligns information across all temporal frames and sample from the 2D features to form a unified 3D feature volume. Finally, this 3D feature volume is fed into the 3D U-Net for further feature extraction and SSC prediction.

In the spatiotemporal module, to spatially align information across all temporal frames, we project all voxels within the current frame's scene range $S$ onto each frame's image plane using the relative poses between past/future frames and the current frame, obtaining projection coordinates $(x_v,y_v,d_v)$.  The visibility of each voxel is then determined by comparing its projected depth $d_v$ with the predicted depth value at the corresponding location in the target frame's depth map $D(x_v,y_v)$ as:
\begin{equation}\label{eq:visible}
V:= \{v|v\in S, |d_v - D(x_v,y_v)| \leq \theta_d\}
\end{equation}
where $\theta_d$ is set to 0.5 meters in experiments to compensate for errors in depth estimation.

To reduce computational cost, we downsample the voxel grid by grouping $4\times4\times4$ voxels into blocks. A block is marked visible if any voxel within it is visible, with projection coordinates averaged over visible voxels. For each visible block, 2D features are sampled from corresponding frames and concatenated (zero-padded if absent) to form the final 3D feature, which is input into the 3D U-Net for SSC.

\definecolor{car}{RGB}{100, 150, 245}
\definecolor{bicycle}{RGB}{100, 230, 245}
\definecolor{motorcycle}{RGB}{30, 60, 150}
\definecolor{truck}{RGB}{80, 30, 180}
\definecolor{other-veh.}{RGB}{100, 80, 250}
\definecolor{person}{RGB}{255, 30, 30}
\definecolor{road}{RGB}{255, 0, 255}
\definecolor{parking}{RGB}{255, 150, 255}
\definecolor{sidewalk}{RGB}{75, 0, 75}
\definecolor{other-grnd.}{RGB}{175, 0, 75}
\definecolor{building}{RGB}{255, 200, 0}
\definecolor{fence}{RGB}{255, 120, 50}
\definecolor{vegetation}{RGB}{0, 175, 0}
\definecolor{terrain}{RGB}{150, 240, 80}
\definecolor{pole}{RGB}{255, 240, 150}
\definecolor{traf.-sign}{RGB}{255, 0, 0}
\definecolor{other-struct.}{RGB}{255, 120, 40}
\definecolor{other-obj.}{RGB}{100, 235, 255}

\begin{table*}[htbp]
\caption{ Quantitative results on the test set of SSCBench-KITTI-360~\cite{li2024sscbench,liao2022kitti}, where the highest and second-highest scores for each metric are highlighted in \textbf{bold} and \underline{underline}, respectively.}
\label{tab:kitti360}
\centering
\footnotesize
\setlength{\tabcolsep}{1pt}
\renewcommand{\arraystretch}{1.2}

\begin{tabular}{c|c|cc|cccccccccccccccccc}
 
\toprule
\textbf{Method} & \textbf{Date} & \textbf{IoU$\uparrow$} & \textbf{mIoU$\uparrow$} & \colorboxsquare{car}{car} & \colorboxsquare{bicycle}{bicycle} & \colorboxsquare{motorcycle}{motorcycle} & \colorboxsquare{truck}{truck} & \colorboxsquare{other-veh.}{other-veh.} & \colorboxsquare{person}{person} & \colorboxsquare{road}{road} & \colorboxsquare{parking}{parking} & \colorboxsquare{sidewalk}{sidewalk} & \colorboxsquare{other-grnd.}{other-grnd.} & \colorboxsquare{building}{building} & \colorboxsquare{fence}{fence} & \colorboxsquare{vegetation}{vegetation} & \colorboxsquare{terrain}{terrain} & \colorboxsquare{pole}{pole} & \colorboxsquare{traf.-sign}{traf.-sign} & \colorboxsquare{other-struct.}{other-struct.} & \colorboxsquare{other-obj.}{other-obj.} \\
\midrule
\multicolumn{21}{@{}l}{\quad\textbf{LiDAR-based methods}} \\ \hline
SSCNet~\cite{song2017semantic} & CVPR2017 & 53.58 & 16.95 & 31.95 & 0.00 & 0.17 & 10.29 & 0.00 & 0.07 & 65.70 & 17.33 & 41.24 & 3.22 & 44.41 & 6.77 & 43.72 & 28.87 & 0.78 & 0.75 & 8.69 & 0.67 \\
LMSCNet~\cite{roldao2020lmscnet}&3DV\phantom{R}2020& 47.35 & 13.65 & 20.91 & 0.00 & 0.00 & 0.26 & 0.58 & 0.00 & 62.95 & 13.51 & 33.51 & 0.20 & 43.67 & 0.33 & 40.01 & 26.80 & 0.00 & 0.00 & 3.63 & 0.00 \\
\hline

\multicolumn{21}{@{}l}{\quad\textbf{Monocular camera-based methods}} \\ \hline
\makebox[2.5cm][c]{MonoScene~\cite{cao2022monoscene}} &CVPR2023& 37.87  & 12.31 & 19.34 & 0.43 & 0.58 & 8.02 & 2.03 & 0.86 & 48.35 & 11.38 & 28.13 & 3.32 & 32.89 & 3.53 & 26.15 & 16.75 & 6.92 & 5.67 & 4.20 & 3.09\\
\makebox[2.5cm][c]{TPVFormer~\cite{huang2023tri}} &CVPR2023& 40.22 & 13.64 & 21.56 & 1.09 & 1.37 & 8.06 & 2.57 & 2.38 & 52.99 & 11.99 & 31.07 & 3.78 & 34.83 & 4.80 & 30.08 & 17.52 & 7.46 & 5.86 & 5.48 & 2.70 \\
\makebox[2.5cm][c]{OccFormer~\cite{zhang2023occformer}} &ICCV2023& 40.27 & 13.81 & 22.58 & 0.66 & 0.26 & 9.89 & 3.82 & 2.77 & 54.30 & 13.44 & 31.53 & 3.55 & 36.42 & 4.80 & 31.00 & 19.51 & 7.77 & 8.51 & 6.95 & 4.60 \\
\makebox[2.5cm][c]{VoxFormer~\cite{li2023voxformer}} &CVPR2023& 38.76 & 11.91 & 17.84 & 1.16 & 0.89 & 4.56 & 2.06 & 1.63 & 47.01 & 9.67 & 27.21 & 2.89 & 31.38 & 4.97 & 28.99 & 14.69 & 6.51 & 6.92 & 3.79 & 2.43 \\
\makebox[2.5cm][c]{IAMSSC~\cite{xiao2024instance}} &T-ITS2024& 41.80 & 12.97 & 18.53 & \underline{2.45} & 1.76 & 5.12 & 3.92 & 3.09 & 47.55 & 10.56 & 28.35 & 4.12 & 31.53 & 6.28 & 29.17 & 15.24 & 8.29 & 7.01 & 6.35 & 4.19 \\
\makebox[2.5cm][c]{DepthSSC~\cite{yao2023depthssc}} &arXiV2024& 40.85 & 14.28 & 21.90 & 2.36 & 4.30 & 11.51 & 4.56 & 2.92 & 50.88 & 12.89 & 30.27 & 2.49 & \underline{37.33} & 5.22 & 29.61 & \underline{21.59} & 5.97 & 7.71 & 5.24 & 3.51 \\
\makebox[2.5cm][c]{Symphonies~\cite{jiang2024symphonize}} &CVPR2024& \underline{44.12} & \underline{18.58} & \textbf{30.02} & 1.85 & \underline{5.90} & \textbf{25.07} & \textbf{12.06} & \textbf{8.20} & \underline{54.94} & \underline{13.83} & \underline{32.76} & \textbf{6.93} & 35.11 & \textbf{8.58} & \textbf{38.33} & 11.52 & \underline{14.01} & \underline{9.57} & \textbf{14.44} & \textbf{11.28}\\

\hline
CF-SSC-Online & - & \textbf{45.79} & \textbf{19.10} & \underline{28.10} & \textbf{3.39} & \textbf{6.87} & \underline{16.76} & \underline{7.75} & \underline{5.68} &\textbf{59.01} & \textbf{16.80} & \textbf{37.60} & \underline{4.95} & \textbf{42.16} & \underline{8.26} & \underline{36.14} & \textbf{21.89} & \textbf{14.73} & \textbf{17.72} & \underline{9.73}& \underline{7.14} \\
\hline
CF-SSC-Offline & - & 47.38 & 19.83 & 28.51 & 3.70 & 6.91 & 17.38 & 7.96 & 6.61 & 62.04 & 17.99 & 39.83 & 4.99 & 42.44 & 8.66 & 38.26 & 24.67 & 15.06 & 18.35 & 9.76& 7.87\\
\bottomrule
\end{tabular}
\end{table*}
\subsection{Losses}
In the CF-SSC framework, since training FutureSynthNet requires supervision in SpatioTemporal SSC's feature space, we decouple the training of SpatioTemporal SSC from other modules. For SpatioTemporal SSC specifically, following \cite{cao2022monoscene}, we employ Scene-Class Affinity Loss $L_{scal}$ with dual supervision on both geometric and semantic aspects, combined with cross-entropy loss weighted by class frequencies. The complete loss function is formulated as:
\begin{equation}\label{eq:loss_ssc}
L = L_{scal}^{geo} + L_{scal}^{sem} +L_{ce}
\end{equation}

For FuturePoseNet, following \cite{wang2017deepvo}, we supervise it using MSE loss. For the future frame images predicted by FutureSynthNet, we apply L1 loss supervision in both image space and feature space, along with SSIM loss for geometric constraints, while using L1 loss to supervise its predicted depth map. The complete loss function is formulated as:

\begin{equation}\label{eq:loss}
L = L_{mse}^{p} + L_{l1}^{img} + L_{l1}^{feat} + L_{ssim}^{img} + L_{l1}^{d}
\end{equation}
The weight for the $L_{mse}^{p}$ loss is set to 0.1, and the weights for all other losses are set to 1.

\section{Experiments}
\subsection{Datasets and Evaluation Metrics}
We conduct experiments on two widely-used real-world traffic scene SSC datasets, SemanticKITTI\cite{behley2019semantickitti} and SSCBench-KITTI-360\cite{li2024sscbench,liao2022kitti}. Both datasets set the scene range to $51.2\text{m}\times51.2\text{m}\times6.4\text{m}$ and define each voxel as a cube with $0.2\text{m}$ edge length, resulting in a $256\times256\times32$ scene resolution. The SemanticKITTI dataset contains 10 training sequences (3,834 samples), 1 validation sequence (815 samples), and 11 hidden test sequences (3,992 samples), with 20 valid classes and 1 invalid class. The SSCBench-KITTI-360 dataset contains 7 training sequences (8,487 samples), 1 validation sequence (1,812 samples), and 1 test sequence (2,566 samples), with 19 valid classes. Consistent with prior SSC tasks, 

Consistent with prior SSC tasks, we employ Intersection over Union (IoU) and mean IoU (mIoU) as evaluation metrics. The IoU metric primarily assesses the model's capability in reconstructing scene geometry, while mIoU evaluates its performance in recovering semantic information of the scene.

\begin{table}[htbp]
\caption{Ablation study on temporal integration methods.}
\begin{center}
\begin{tabular}{c|cc}
\toprule
\textbf{Methods}  & \textbf{IoU$\uparrow$} & \textbf{mIoU$\uparrow$} \\
\hline
\textbf{Concatenation} & 44.5 & 14.7 \\
\hline
\textbf{Spatiotemporal Module} & 48.6 (+4.1) & 17.0 (+2.3) \\
\bottomrule
\end{tabular}
\label{tabint}
\end{center}
\end{table}
\begin{table}[htbp]
\caption{Ablation study for temporal input on SpatioTemporal SSC.}
\begin{center}
\begin{tabular}{cccccccccc}
\toprule
\textbf{t} & -20 & -15 & -10 &-5 & 0 & +5 & +5p & \textbf{IoU$\uparrow$} & \textbf{mIoU$\uparrow$} \\
\midrule
\multirow{2}{*}{\textbf{Offline}} &  & &  &  & \checkmark & \checkmark &  & 46.2 & 15.7 \\
 & \checkmark &\checkmark & \checkmark & \checkmark & \checkmark & \checkmark &  & 48.6 & 17.0 \\
\hline
\multirow{5}{*}{\textbf{Online}} &  & &  &  & \checkmark &  &  & 43.7 & 14.3 \\
&   &  &  \checkmark & \checkmark  & \checkmark &   &  & 45.8 & 14.8 \\

&  \checkmark & \checkmark &  \checkmark & \checkmark  & \checkmark &   &  & 46.3 & 14.9 \\

& \checkmark &\checkmark & \checkmark & \checkmark & \checkmark &  & \checkmark & 46.8 & 16.0 \\
&  &  &  &  & \checkmark &  & \checkmark & 44.3 & 15.2 \\
\bottomrule
\end{tabular}
\label{tabt}
\end{center}
\end{table}
\begin{figure*}[htbp]
  \centering
\includegraphics[width=0.8\linewidth]{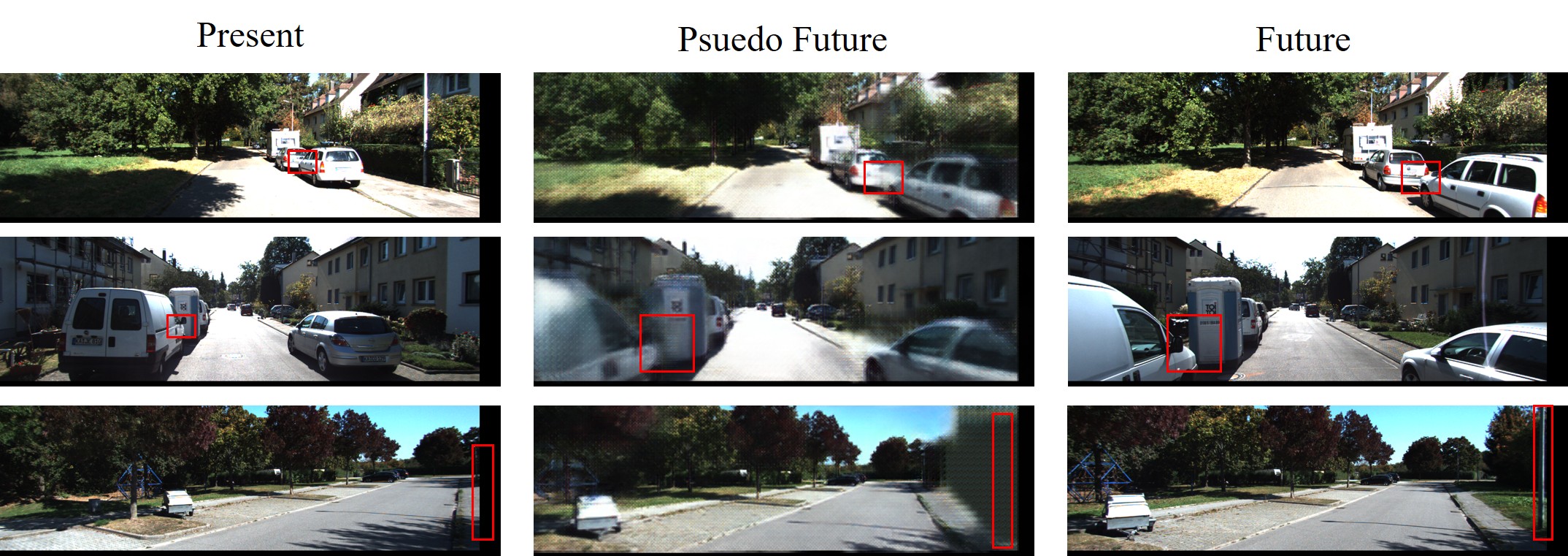}
   \caption{Visualization of occlusion handling.}
   \label{fig:occ}
\end{figure*}
\subsection{Implementation Details}
We train our model using 4 NVIDIA RTX 4090 GPUs. Following previous methods\cite{cao2022monoscene}, we set the batch size to 4 and employ the AdamW\cite{loshchilov2017decoupled} optimizer for 30 epochs. The initial learning rate is set to 1e-4 with weight decay of 1e-4, and the learning rate is reduced by a factor of 0.1 at the 20th and 25th epochs. 

Both SpatioTemporal SSC and FuturePoseNet use ResNet-50\cite{he2016deep} as the backbone, initialized with MaskDINO-pretrained weights\cite{li2023mask}, following \cite{jiang2024symphonize}. The 3D U-Net, based on \cite{cao2022monoscene, roldao2020lmscnet}, applies two downsampling and upsampling stages, with each $3\times3\times3$ convolution decomposed into three $3\times1\times1$ convolutions to reduce parameters. For depth estimation, we adopt the off-the-shelf method from \cite{li2023voxformer}, a standard choice in prior works\cite{jiang2024symphonize,zheng2024monoocc,wang2024h2gformer,wang2024not,yao2023depthssc}.


Additionally, as mentioned earlier, we use a frame interval of 5 frames between consecutive inputs

\subsection{Main Results}
We conducted comparisons with a series of state-of-the-art (SOTA) methods on the SemanticKITTI\cite{behley2019semantickitti} and SSCBench-KITTI-360\cite{li2024sscbench,liao2022kitti} datasets, as shown in Tab.\ref{tab:smkitti} and Tab.\ref{tab:kitti360}. Here, \textbf{offline} methods utilize ground-truth future frame information and can only operate in offline mode. These are included in the table for reference but are not directly compared with other methods. In contrast, \textbf{online} methods rely solely on the current and past frames and are compared against other approaches.

As shown in the results, our CF-SSC (online version) achieves 16.4\% mIoU on the SemanticKITTI dataset, surpassing all existing monocular SSC methods. Notably, our approach even outperforms StereoScene\cite{li2023bridging}, a novel stereo SSC method. On the SSCBench-KITTI-360 dataset, we also achieve 19.1\% mIoU, exceeding the current SOTA methods. These results demonstrate that "seeing ahead" significantly enhances monocular SSC performance.

Finally, We provide a visual comparison between our SSC predictions and MonoScene\cite{cao2022monoscene} on the SemanticKITTI dataset, as illustrated in the Fig.\ref{fig:vis}. Our method demonstrates superior object recognition and scene reconstruction.

As shown in the Fig.\ref{fig:occ}, we use L1 loss on primary features to suppress artifacts, resulting in blurry outputs. While mildly occluded regions are reconstructed well, severe occlusions remain challenging.

\subsection{Ablations}
We conduct comprehensive ablation studies on the SemanticKITTI\cite{behley2019semantickitti} validation set to evaluate: (1) the impact of different temporal integration methods on network performance, (2) the effect of temporal frame inputs on SpatioTemporal SSC, and the performance gains from incorporating pseudo future frames, pseudo depth maps, and predicted future frame poses.

\noindent$\mathbf{The \ Role \ of \ SpatialTemporal \ Module.}$ As shown in Tab.\ref{tabint}, we compare different temporal integration methods for SSC, where both models take 4 past frames and 1 future frame as additional inputs with a 5-frame interval. The baseline "concatenation" method (commonly adopted in previous works) directly concatenates 2D features from different frames with the current frame's 2D features before lifting them to 3D space using the current frame's predicted depth for SSC. Our SpatioTemporal Module significantly improves both IoU and mIoU metrics, demonstrating its effectiveness and highlighting the benefits of temporal feature integration in 3D space rather than 2D space.

\noindent$\mathbf{Different \ Temporal \ Input.}$ We conduct a comprehensive analysis of different temporal input configurations' impact on model performance (as shown in Tab.\ref{tabt}), where $\mathbf{+5p}$ represents using our predicted pseudo future frames, depth maps, and poses. Optimal results require ground-truth future frames, which are unavailable online. In practical settings, current-frame-only input performs worst; adding past frames yields limited gains due to viewpoint misalignment. Our framework significantly improves performance by integrating FutureSynthNet-generated pseudo future frames and depth maps with FuturePoseNet-predicted poses.


\section{Conclusion}
In this paper, we present CF-SSC, a temporal SSC framework that tackles monocular SSC's limited observation range by predicting future frames from past observations. Experiments show SOTA performance on SemanticKITTI and SSCBench-KITTI-360.We hope this work inspires future research in temporal 3D scene understanding, advancing monocular perception in dynamic environments.

\bibliographystyle{IEEEtran}
\bibliography{IEEEexample}

\end{document}